\documentclass{article}

\usepackage{microtype}
\usepackage{graphicx}
\usepackage{subfigure}
\usepackage[dvipsnames]{xcolor}
\usepackage{booktabs} 
\usepackage{bm}
\usepackage{hyperref}
\usepackage{multirow}
\usepackage{multicol}
\usepackage{colortbl}
\usepackage{tabulary}
\usepackage{etoolbox}
\definecolor{mycolor_blue}{HTML}{E7EFFA}
\definecolor{mycolor_green}{HTML}{E6F8E0}
\definecolor{mycolor_gray}{HTML}{ECECEC}
\definecolor{pearDark}{HTML}{2980B9}

\usepackage[accepted]{icml2024}

\usepackage{amsmath}
\usepackage{amssymb}
\usepackage{mathtools}
\usepackage{amsthm}
\usepackage{tcolorbox}
\usepackage[capitalize,noabbrev]{cleveref}

\def\vx{{\bm{x}}}

\def\vz{{\bm{z}}}

\theoremstyle{plain}

\theoremstyle{definition}

\theoremstyle{remark}

\usepackage[textsize=tiny]{todonotes}

\icmltitlerunning{Mastering Text-to-Image Diffusion: Recaptioning, Planning, and Generating with Multimodal LLMs}
\begin{document}

\twocolumn[
\icmltitle{Mastering Text-to-Image Diffusion: \\Recaptioning, Planning, and Generating with Multimodal LLMs}


\icmlsetsymbol{equal}{*}

\begin{icmlauthorlist}
\icmlauthor{Ling Yang}{equal,yyy}
\icmlauthor{Zhaochen Yu}{equal,yyy}
\icmlauthor{Chenlin Meng}{comp,comp2}
\icmlauthor{Minkai Xu}{comp}
\icmlauthor{Stefano Ermon}{comp}
\icmlauthor{Bin Cui}{yyy}
\url{https://github.com/YangLing0818/RPG-DiffusionMaster}
\end{icmlauthorlist}

\icmlaffiliation{yyy}{Peking University, China}
\icmlaffiliation{comp}{Stanford University, USA}
\icmlaffiliation{comp2}{Pika Labs, USA}
\icmlcorrespondingauthor{Ling Yang}{yangling0818@163.com}

\icmlkeywords{Machine Learning, ICML}

\vskip 0.3in
]



\printAffiliationsAndNotice{\icmlEqualContribution} 

\begin{abstract}
\vspace{-0.2in}
Diffusion models have exhibit exceptional performance in text-to-image generation and editing. However, existing methods often face challenges when handling complex text prompts that involve multiple objects with multiple attributes and relationships. In this paper, we propose a brand new \textit{training-free} text-to-image generation/editing framework, namely \textbf{\textit{Recaption, Plan and Generate (RPG)}}, harnessing the powerful chain-of-thought reasoning ability of multimodal LLMs to enhance the compositionality of text-to-image diffusion models. Our approach employs the MLLM as a global planner to decompose the process of generating complex images into multiple simpler generation tasks within subregions. We propose \textit{complementary regional diffusion} to enable region-wise compositional generation. Furthermore, we integrate text-guided image generation and editing within the proposed RPG in a closed-loop fashion, thereby enhancing generalization ability. Extensive experiments demonstrate our RPG outperforms state-of-the-art text-to-image diffusion models, including DALL-E 3 and SDXL, particularly in multi-category object composition and text-image semantic alignment. Notably, our RPG framework exhibits wide compatibility with various MLLM architectures (e.g., MiniGPT-4) and diffusion backbones (e.g., ControlNet). Our code is available at \href{https://github.com/YangLing0818/RPG-DiffusionMaster}{https://github.com/YangLing0818/RPG-DiffusionMaster}
\end{abstract}

\section{Introduction}
Recent advancements in diffusion models \citep{sohl2015deep,dhariwal2021diffusion,song2020score,yang2023diffusion} have significantly improve the synthesis results of text-to-image models, such as Imagen \citep{saharia2022photorealistic}, DALL-E 2/3 \citep{ramesh2022hierarchical,betker2023improving} and SDXL \citep{podell2023sdxl}. However, despite their remarkable capabilities in synthesizing realistic images consistent with text prompts, most diffusion models usually struggle to accurately follow some complex prompts \citep{feng2022training,lian2023llm,liu2022compositional,bar2023multidiffusion}, which require the model to compose objects
with different attributes and relationships into a single image.

\begin{figure}[t]
\vspace{-0.2in}
\begin{center}\centerline{\includegraphics[width=1.\linewidth]{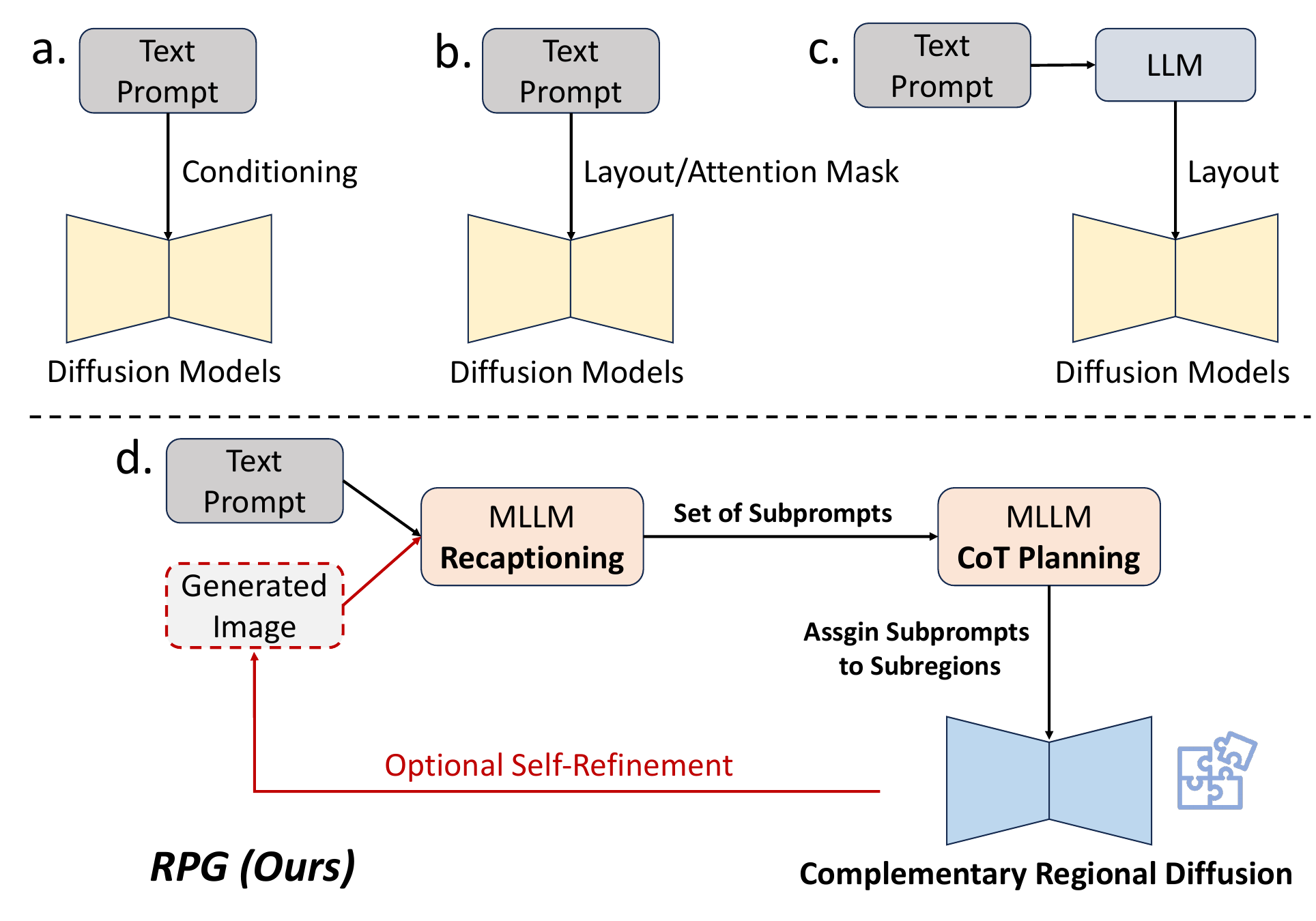}}
\vspace{-0.2in}
\caption{Architecture comparison between (a) text-conditional diffusion models \citep{ramesh2022hierarchical}, (b) layout/attention-based diffusion models \citep{feng2022training,cao2023masactrl}, (c) LLM-grounded diffusion models \citep{lian2023llm} and (d) our RPG.}
\label{pic-RPG-intro}
\end{center}
\vspace{-0.4in}
\end{figure}

Some works introduce additional layouts/boxes \citep{li2023gligen,xie2023boxdiff,yang2023reco,qu2023layoutllm,chen2024training,wu2023self,lian2023llm} as conditions or leveraging prompt-aware attention guidance \citep{feng2022training,chefer2023attend,wang2023compositional} to improve compositional text-to-image synthesis. For example, StructureDiffusion \citep{feng2022training} incorporates linguistic structures into the guided generation process by manipulating cross-attention maps in diffusion models. GLIGEN \citep{li2023gligen} designs trainable gated self-attention layers to incorporate spatial inputs,
such as bounding boxes, while freezing the weights of original diffusion model. 

\begin{figure*}
\begin{center}\centerline{\includegraphics[width=0.8\linewidth]{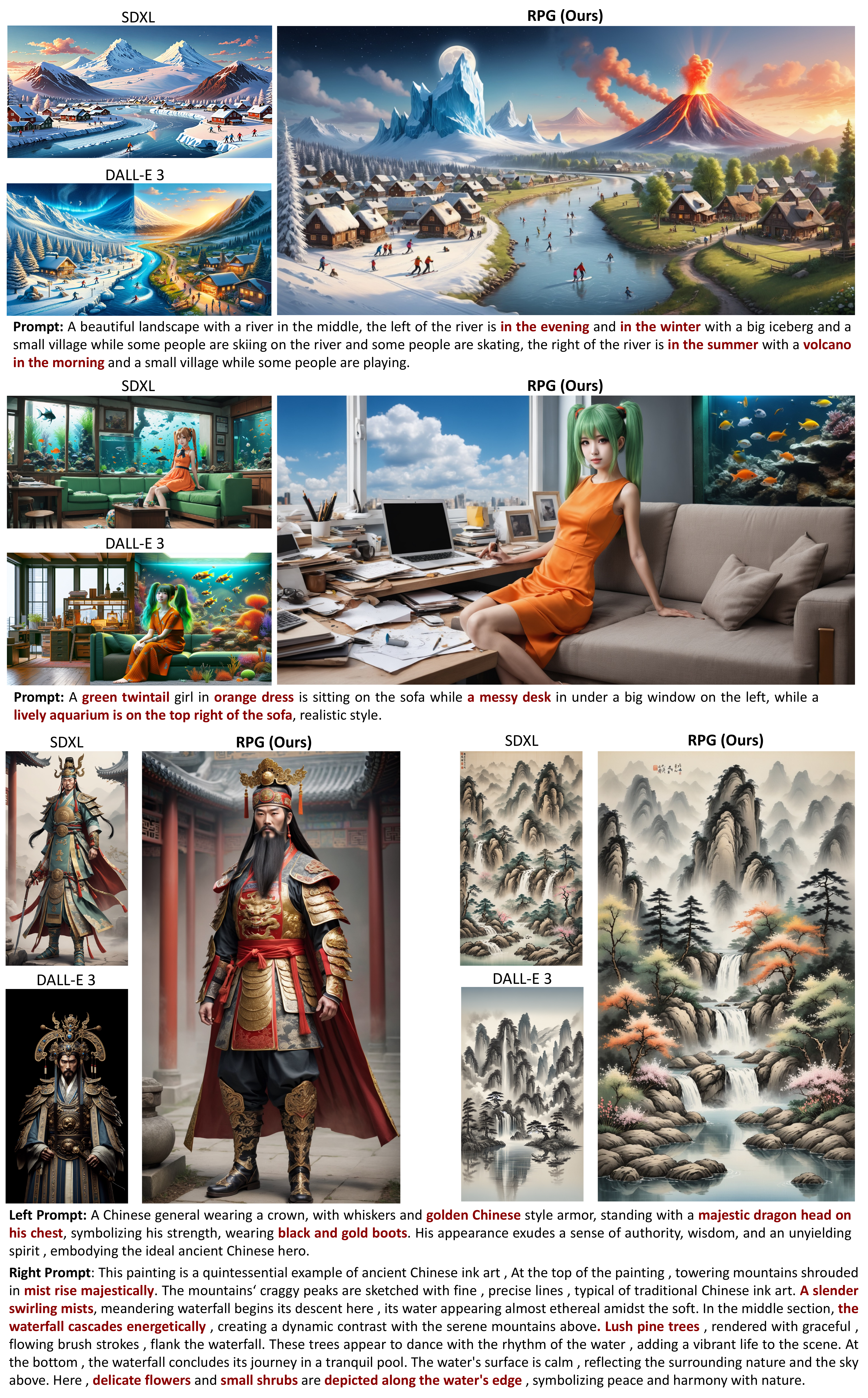}}
\caption{Compared to SDXL \citep{podell2023sdxl} and DALL-E 3 \citep{betker2023improving}, our proposed RPG exhibits a superior ability to convey intricate and compositional text prompts within generated images (\textbf{colored text denotes critical part}).}
\label{pic-RPG-qualitative-first}
\end{center}
\vspace{-7mm}
\end{figure*}

\begin{figure*}
\begin{center}\centerline{\includegraphics[width=1.\linewidth]{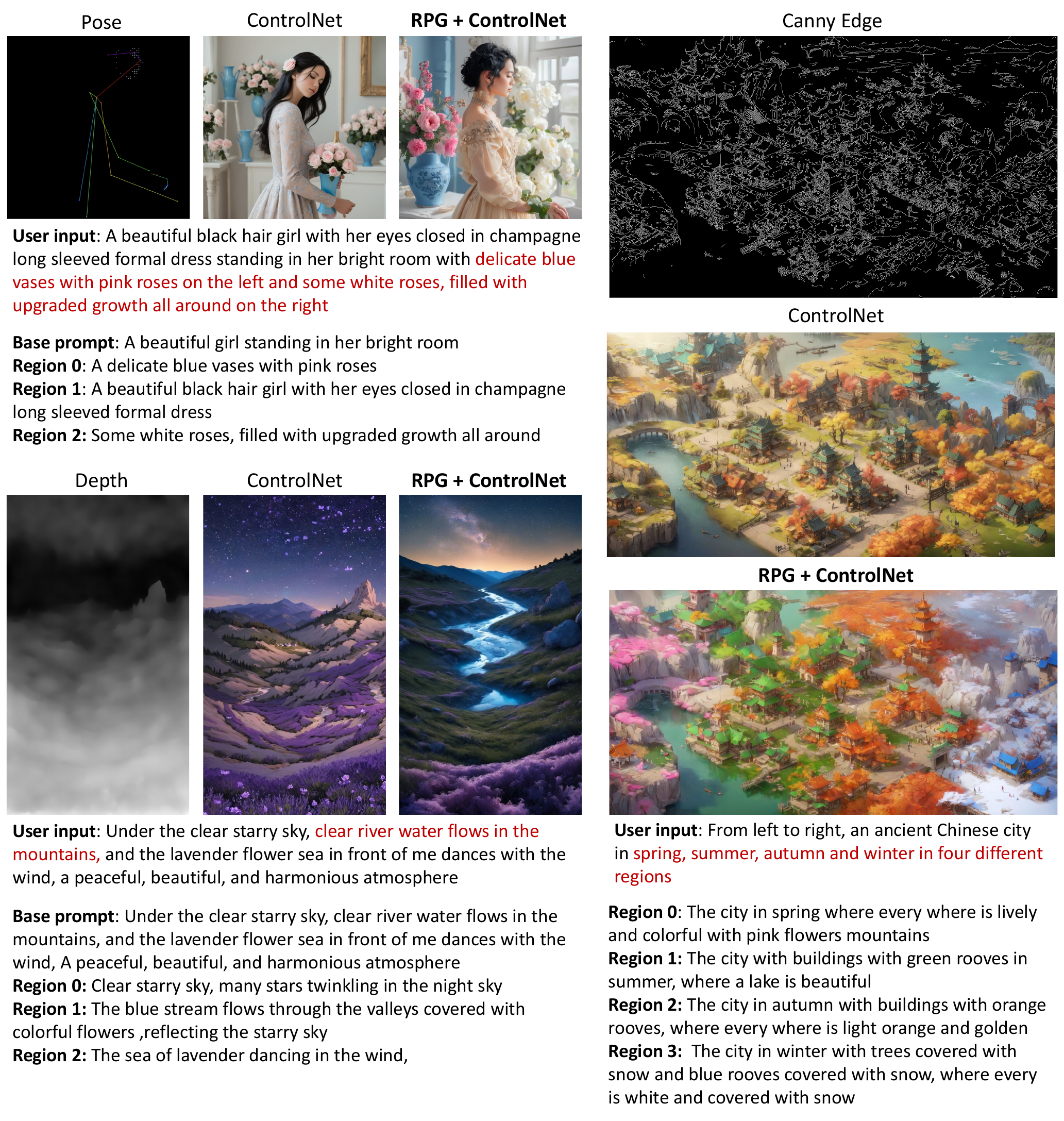}}
\caption{Our RPG framework can extend text-to-image generation with more conditions (e.g., pose, depth and canny edge) by utilizing ControlNet \citep{zhang2023adding}. Compared to original ControlNet, RPG significantly improves its prompt understanding by decomposing "user input" into the combination of base prompt and subprompts, and further enhance its compositional semantic alignment of generated images by performing region-wise diffusion generation (in \cref{sec-generation}).}
\label{pic-RPG-qualitative-controlnet}
\end{center}
\vspace{-7mm}
\end{figure*}

Another potential solution is to leverage 
image understanding feedback \citep{huang2023t2i,xu2023imagereward,sun2023dreamsync,fang2023boosting}  for refining diffusion generation. For instance, GORS \citep{huang2023t2i} finetunes a pretrained text-to-image model with generated images that highly align with the compositional prompts, where the fine-tuning loss is weighted by the text-image alignment reward.
Inspired by the reinforcement learning from human
feedback (RLHF) \citep{ouyang2022training,stiennon2020learning} in natural language processing, ImageReward \citep{xu2023imagereward} builds a general-purpose reward model to improve text-to-image models in aligning with human preference.

Despite some improvements achieved by these methods, there are still two main limitations in the context of compositional/complex image generation: (i) existing layout-based or attention-based methods can only provide rough and suboptimal spatial guidance, and struggle to deal with overlapped objects \citep{cao2023masactrl,hertz2022prompt,lian2023llm} ; (ii) feedback-based methods require to collect high-quality feedback and incur additional training costs.

To address these limitations, we introduce a new \textit{training-free} text-to-image generation framework, namely \textbf{\textit{Recaption, Plan and Generate (RPG)}}, unleashing the impressive reasoning ability of multimodal LLMs to enhance the compositionality and controllability of diffusion models. We propose three core strategies in RPG:

\textbf{Multimodal Recaptioning.} We specialize in transforming text prompts into highly descriptive ones, offering informative augmented prompt comprehension and semantic alignment in diffusion models.
We use LLMs to decompose the text prompt into distinct subprompts, and recaption them with more detailed descriptions. We use MLLMs to automatically recaption input image for identifying the semantic discrepancies between generated images and target prompt.

\textbf{Chain-of-Thought Planning.}  In a pioneering approach, we partition the image space into complementary subregions and assign different subprompts to each subregion, breaking down compositional generation tasks into multiple simpler subtasks. Thoughtfully crafting task instructions and in-context examples, we harness the powerful chain-of-thought reasoning capabilities of MLLMs \citep{zhang2023multimodal} for efficient region division. By analyzing the recaptioned intermediate results, we generate detailed rationales and precise instructions for subsequent image compositions.

\begin{figure*}[t]
\begin{center}\centerline{\includegraphics[width=1\linewidth]{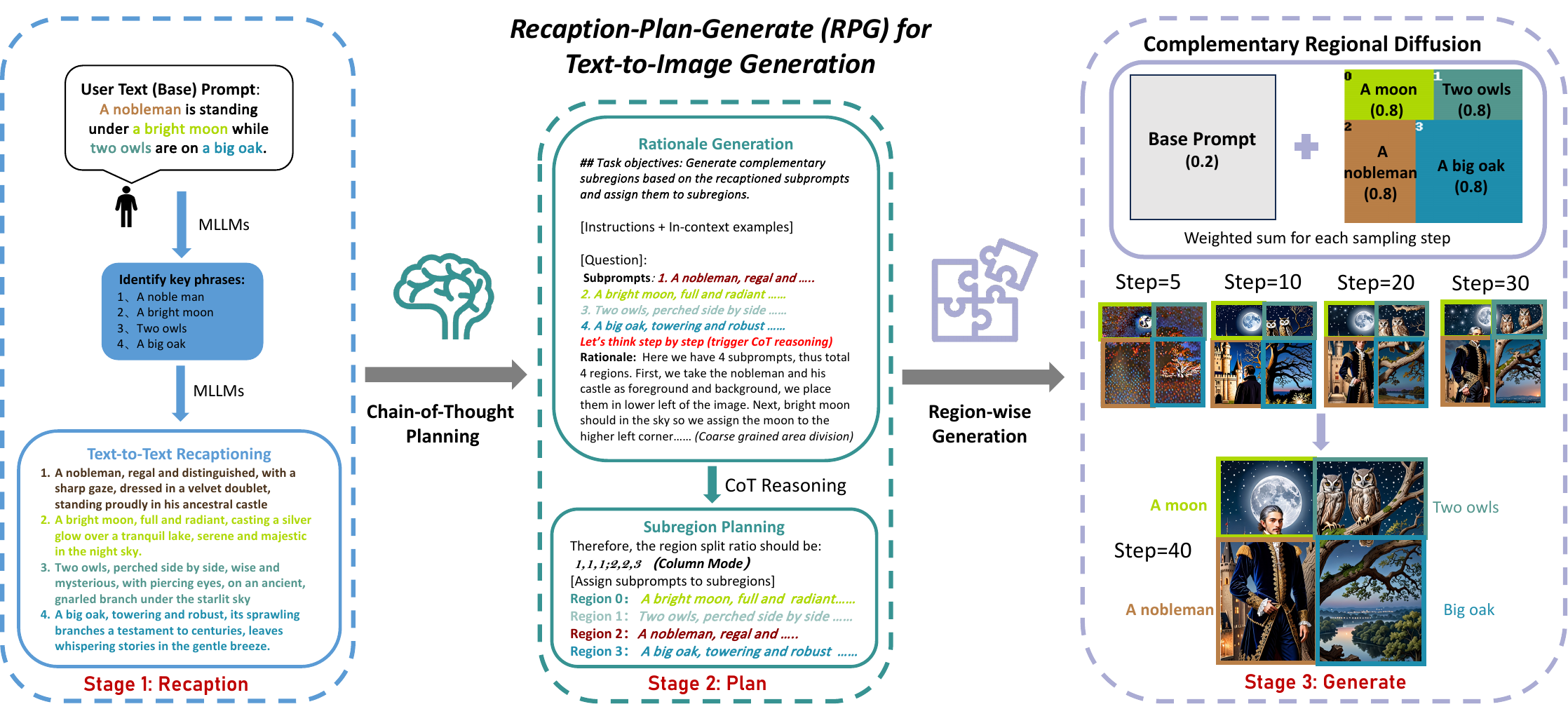}}
\caption{Overview of our RPG framework for text-to-image generation.}
\label{pic-RPG-generation}
\end{center}
\vspace{-0.4in}
\end{figure*}

\textbf{Complementary Regional Diffusion.}  Based on the planned non-overlapping subregions and their respective prompts, we propose \textit{complementary regional diffusion} to enhance the flexibility and precision of compositional text-to-image generation. Specifically, we independently generate image content guided by subprompts within designated rectangle subregion, and subsequently merge them spatially in a \textit{resize-and-concatenate} approach. This region-specific diffusion effectively addresses the challenge of conflicting overlapped image contents. 
Furthermore, we extend this framework to accommodate editing tasks by employing contour-based regional diffusion, enabling precise manipulation of inconsistent regions targeted for modification.  

This new RPG framework can unify both text-guided image generation and editing tasks in a closed-loop fashion. We compare our RPG framework with previous work in \cref{pic-RPG-intro} and summarize our main contributions as follows:

\begin{itemize}
    \item We propose a new training-free text-to-image generation framework, namely \textit{Recaption, Plan and Generate (RPG)}, to improve the composibility and controllability of diffusion models to the fullest extent.
    \item RPG is the first to utilize MLLMs as both \textit{multimodal recaptioner and CoT planner} to reason out more informative instructions for steering diffusion models.
    \item We propose \textit{complementary regional diffusion} to enable extreme collaboration with MLLMs for compositional image generation and precise image editing.  
    \item Our RPG framework is user-friendly, and can be generalized to different MLLM architectures (e.g., MiniGPT-4) and diffusion backbones (e.g., ControlNet). 
    \item Extensive qualitative and quantitative comparisons with previous SOTA methods, such as SDXL, DALL-E 3 and InstructPix2Pix, demonstrate our superior text-guided image generation/editing ability.
\end{itemize}

\section{Method}
\subsection{Overview of Proposed RPG}
\label{sec-framework}
In this section, we introduce our novel training-free framework - \textbf{R}ecaption, \textbf{P}lan and \textbf{G}enerate (\textbf{RPG}). 
We delineate three fundamental strategies of our RPG in text-to-image generation (\cref{sec-generation}),  as depicted in \cref{pic-RPG-generation}. Specifically, given a complex text prompt that includes multiple entities and relationships, we leverage (multimodal) LLMs to \textit{recaption} the prompt by decomposing it into a base prompt and highly descriptive subprompts. Subsequently, we utilize multimodal CoT planning to allocate the split (sub)prompts to complementary regions along the spatial axes. Building upon these assignments, we introduce \textit{complementary regional diffusion} to independently generate image latents and aggregate them in each sampling step.

Our RPG framework exhibits versatility by extending its application to text-guided image editing with minimal adjustments, as exemplified in \cref{sec-editing}. For instance, in the recaptioning phase, we utilize MLLMs to analyze the paired target prompt and source image, which results in informative multimodal feedback that captures their cross-modal semantic discrepancies. In multimodal CoT planning,  we generate a step-by-step edit plan and produce \textit{precise contours} for our regional diffusion. Furthermore, we demonstrate the ability to execute our RPG workflow in a closed-loop manner for progressive self-refinement, as showcased in \cref{sec-editing}. This approach combines precise contour-based editing with complementary regional diffusion generation.

\subsection{Text-to-image Generation}
\label{sec-generation}
\paragraph{Prompt Recaptioning}
Let $y^c $ be a complex user prompt which includes multiple entities with different attributes and relationships. We use MLLMs to identify the key phrases in $y^c$ to obtain subpormpts denoted as:
\begin{equation}
\label{eq-1}
\{y^i\}_{i=0}^n = \{y^0,y^1,...,y^n\} \subseteq y^c,
\end{equation}
where $n$ denotes the number of key phrases.  Inspired by DALL-E 3 \citep{betker2023improving}, which uses pre-trained \textbf{image-to-text} (I2T) caption models to generate descriptive prompts for images, and construct new datasets with high-quality image-text pairs. In contrast, we leverage the impressive language understanding and reasoning abilities of LLMs and use the LLM as the \textbf{text-to-text} (T2T) captioner to further \textit{recaption} each subprompt with more informative detailed descriptions:
\begin{equation}
\label{eq-2}
 \{\hat{y}^0,\hat{y}^1,...,\hat{y}^n\} = \text{Recaption}(\{{y^i}\}_{i=0}^n).
\end{equation} 
In this way, we can produce denser fine-grained details for each subprompt in order to effectively improve the fidelity of generated image, and reduce the semantic discrepancy between prompt and image.
\begin{figure}[h]
\begin{center}\centerline{\includegraphics[width=1.\linewidth]{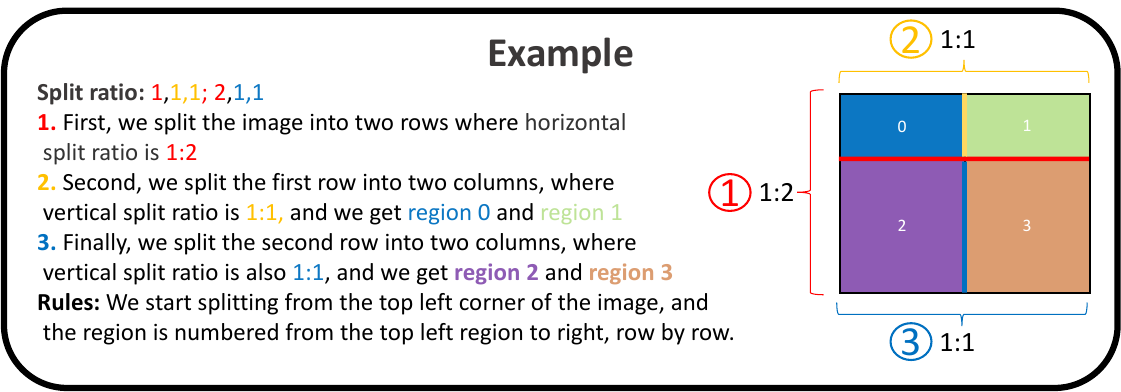}}
\caption{An illustrative example for region division.}
\label{pic-splitratio}
\end{center}
\end{figure}

\begin{figure*}[t]
\begin{center}\centerline{\includegraphics[width=1\linewidth]{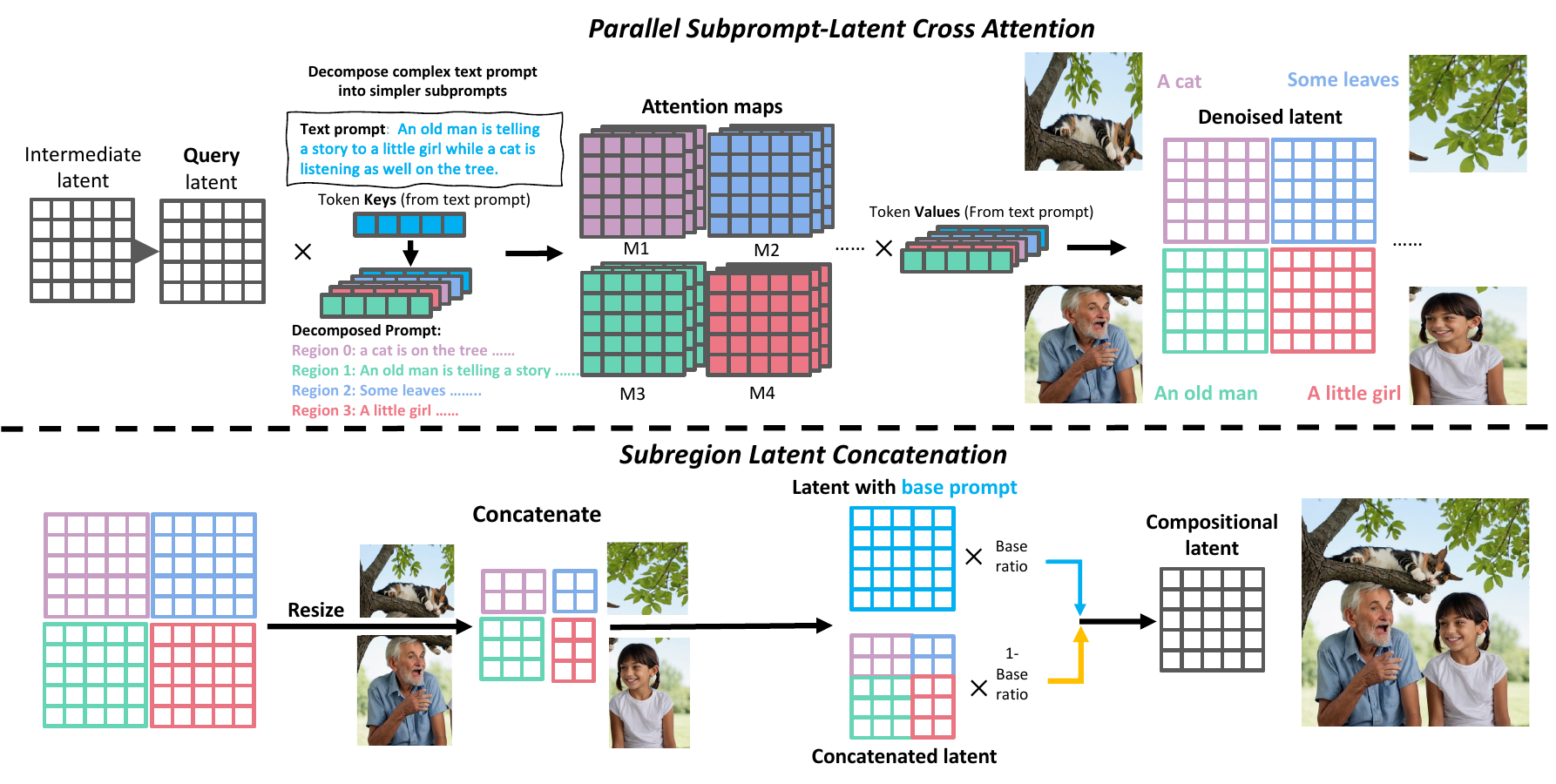}}
\caption{The demonstration of each sampling step in our \textbf{Complementary Regional Diffusion}.}
\label{pic-regional}
\end{center}
\end{figure*}

\paragraph{CoT Planning for Region Division}
Based on the recaptioned subprompts, we leverage the powerful multimodal chain-of-thought (CoT) reasoning ability of LLMs \citep{zhang2023multimodal} to plan the compositions of final image content for diffusion models. Concretely, we divide image space ${H\times W}$ into several \textit{complementary regions}, and assign each augmented subprompt $\hat{y}^i$ to specific region $R^i$:
\begin{equation}\label{eq-3}
    \{R^i\}_{i=0}^n =\{R^0,R^1,...,R^n\} \subseteq{H\times W},
\end{equation} 
In order to produce meaningful and accurate subregions, we need to carefully specify two components for planning region divisions: (i) region parameters: we define that rows are separated by ";" and each column is denoted by a series of numbers separated by commas (e.g., "1,1,1"). 
To be specific , we first use ";" to split an image into different rows, then within each row, we use commas to split a row into different regions, see \cref{pic-splitratio} for better comprehension; (ii) region-wise task specifications to instruct MLLMs: we utilize the CoT reasoning of MLLMs with some designed in-context examples to reason out the plan of region division. We here provide a simplified template of our instructions and in-context examples:
\begin{tcolorbox}  
{\slshape 
\textcolor{NavyBlue}{\textbf{1.Task instruction}}\\
You are an smart region planner for image. You should use split ratio to specify the split method of the image, and then recaption each subregion prompts with more descriptive prompts while maintaining the original meaning.\\
\textcolor{BlueGreen}{\textbf{2.Multi-modal split tutorial
}}\\
\centerline{......}\\
\textcolor{Plum}{\textbf{3. In-context examples}}\\
\textbf{User Prompt:} A girl with white ponytail and black dress are chatting with a blonde curly hair girl in a white dress in a cafe.\\
\# Key pharses extraction and Recaption\\
\# Split ratio Planning\\
\# Composition Logic\\
\# Aesthetic Considerations:\\
\# Final output\\
\textcolor{ForestGreen}{\textbf{4.Trigger CoT reasoning ability of MLLMs}}\\
\textbf{User Prompt:} An old man with his dog is looking at a parrot on the tree.\\
Reasoning: Let's think step by step......
}
\end{tcolorbox}
To facilitating inferring the region for each subprompt, we adhere to two key principles in designing in-context example and generating informative rationales: (i) the objects with same class name (e.g., five apples) will be separately assign to different regions to ensure the numeric accuracy; (ii) If the prompt focuses more on the appearance of a specific entity, we treat the different parts of this entity as different entities (e.g., A green hair twintail in red blouse , wearing blue skirt. $\Longrightarrow$ green hair twintail, red blouse, blue skirt).  

\begin{figure*}[t]
\begin{center}\centerline{\includegraphics[width=1\linewidth]{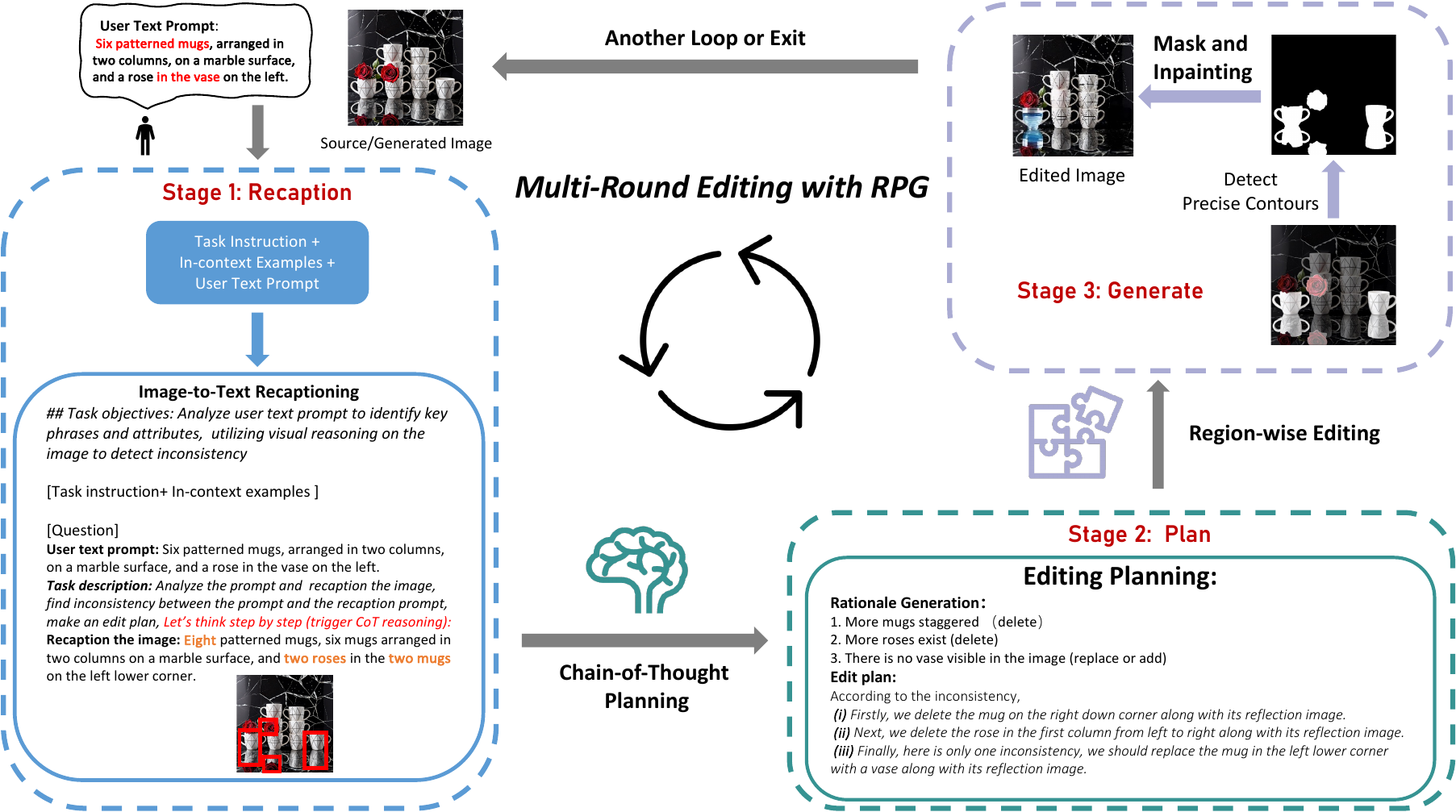}}
\caption{RPG unifies text-guided image generation and editing in a closed-loop approach.}
\label{pic-RPG-editing}
\end{center}
\end{figure*}

\paragraph{Complementary Regional Diffusion}
Recent works \citep{liu2022compositional,wang2023compositional,chefer2023attend,feng2022training} have adjusted cross-attention masks or layouts to facilitate compositional generation. However, these approaches predominantly rely on simply stacking latents, leading to conflicts and ambiguous results in overlapped regions. To address this issue, as depicted in \cref{pic-regional}, we introduce a novel approach called complementary regional diffusion for region-wise generation and image composition. We extract non-overlapping complementary rectangular regions and apply a \textbf{resize-and-concatenate} post-processing step to achieve high-quality compositional generation. Additionally, we enhance coherence by combining the base prompt with recaptioned subprompts to reinforce the conjunction of each generated region and maintain overall image coherence (detailed ablation study in \cref{sec-abaltion}). This can be represented as:
 \begin{equation}
     \label{eq-4}
     \vx_{t-1}= \text{CRD}(\vx_t,y^\text{\text{base}},\{\hat{y}^i\}_{i=0}^n,\{R^i\}_{i=0}^n,t,s),
 \end{equation} 
where $s$ is a fixed random seed, $\text{CRD}$ is the abbreviation for complementary regional diffusion. 

More concretely, we construct a prompt batch with base prompt $y^{\text{base}}=y^c$  and the recaptioned subprompts:
\begin{equation}
\label{eq-5}
\text{Prompt Batch:}\quad \{y^\text{\text{base}},\{\hat{y}^i\}_{i=0}^n\}.
\end{equation} 
In each timestep, we deliver the prompt batch into the denoising network and manipulate the cross-attention layers to generate different latents $\{\vz^i_{t-1}\}_{i=0}^n$ and $\vz_{t-1}^\text{base}$ in \textit{parallel}, as demonstrated in \cref{pic-regional}. We formulate this process as: 
\begin{equation}
\label{eq-6}
\vz_{t-1}^i=\text{Softmax}(\frac{(W_Q \cdot \mathcal{\phi}(\vz_t))(W_K \cdot \mathcal{\psi}(\hat
y^i))^T}{\sqrt{d}})W_V\cdot\mathcal{\psi}(\hat
y^i),
\end{equation}
where image latent $\vz_t$ is the query and each subprompt $\hat y^i$ works as a key and value. $W_Q ,W_K, W_V$ are linear projections and $d$ is the latent projection dimension of the keys and queries.
Then, we shall proceed with resizing and concatenating the generated latents $\{\vz_{t-1}^i\}^n_{i=0}$, according to their assigned region numbers (from $0$ to $n$) and respective proportions. Here we denote each resized latent as:
\begin{equation}
\label{eq-7}
\vz_{t-1}^i(h,w)=\text{Resize}(\vz_{t-1}^i,R^i),
\end{equation} 
where $h,w$ are the height and the width of its assigned region $R^i$. We directly concatenate them along the spatial axes: 
\begin{equation}
\label{eq-8}
    \vz_{t-1}^\text{cat}= \text{Concatenate}(\{\vz_{t-1}^i(h,w)\}^n_{i=0}).
\end{equation} 
To ensure a coherent transition in the boundaries of different regions and a harmonious fusion between the background and the entities within each region, we use the weighted sum of the \textit{base latents} $\vz_{t-1}^\text{\text{base}}$ and the \textit{concatenated latent} $\vz_{t-1}^\text{cat}$ to produce the final denoising output:
\begin{equation}
\label{eq-9}
    \vz_{t-1}=\beta * \vz_{t-1}^{\text{\text{base}}} + (1-\beta) * \vz_{t-1}^{\text{cat}}.
\end{equation}
Here $\beta$ is used to achieve a suitable balance between human aesthetic perception and alignment with the complex text prompt of the generated image. It is worth noting that complementary regional diffusion can generalize to arbitrary diffusion backbones including SDXL \citep{podell2023sdxl}, ConPreDiff \citep{yang2023improving} and ControlNet \citep{zhang2023adding}, which will be evaluated in \cref{sec-exp-generation}. 

\begin{figure*}[ht]
\begin{center}\centerline{\includegraphics[width=1.0\linewidth]{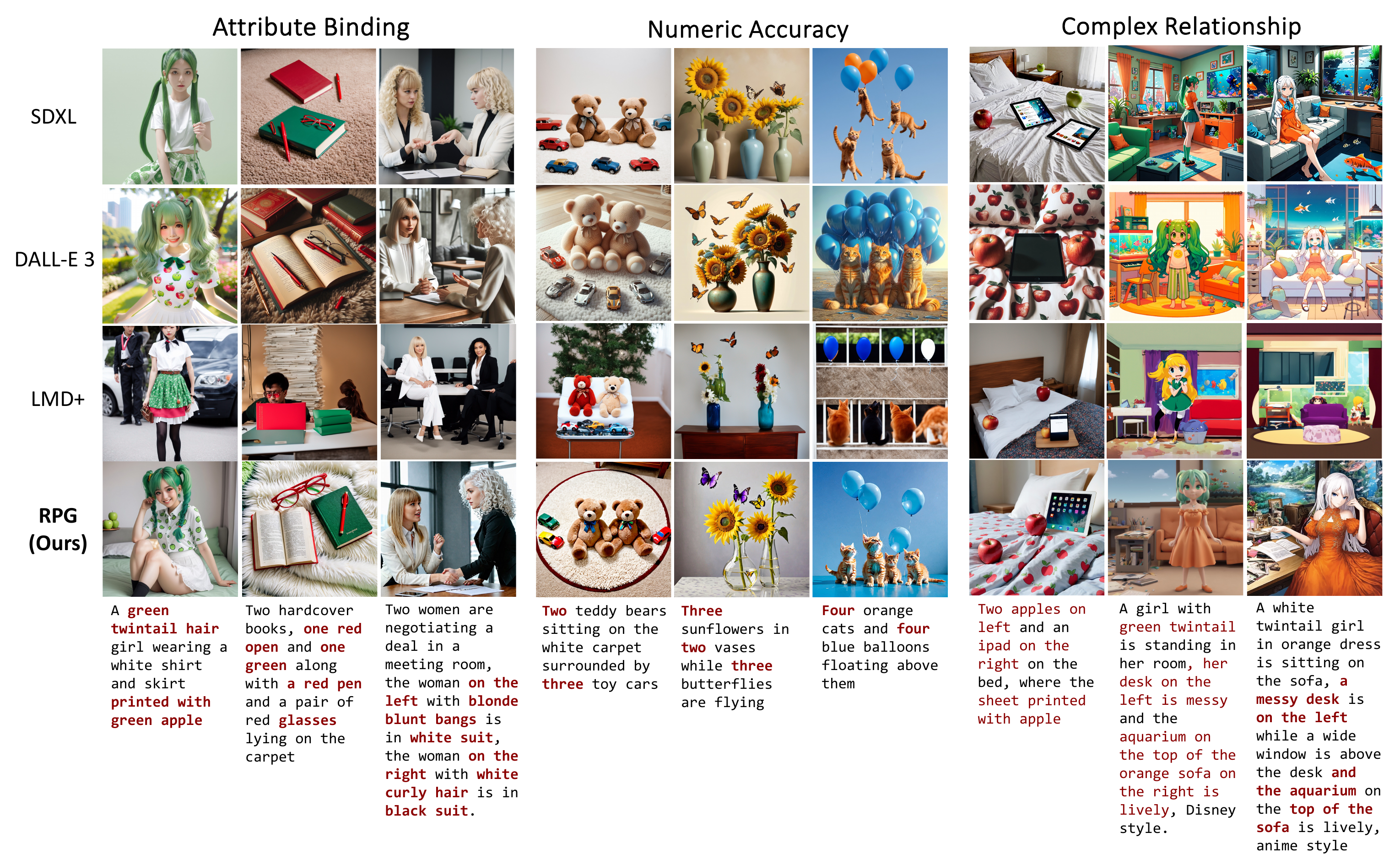}}
\vspace{-0.2in}
\caption{Qualitative comparison between our RPG and SOTA text-to-image models (SDXL \citep{podell2023sdxl} and DALL-E 3 \citep{betker2023improving}), and LLM-grounded diffusion model LMD+ \citep{lian2023llm}.}
\label{pic-RPG-main}
\end{center}
\vspace{-0.3in}
\end{figure*}

\subsection{Text-Guided Image Editing}
\label{sec-editing}
\paragraph{Image Recaptioning}
Our RPG can also generalize to text-guided image editing tasks as illustrated in 
\cref{pic-RPG-editing}. In recaptioning stage, RPG adopts MLLMs  as a captioner to recaption the source image, and leverage its powerful reasoning ability to identify the fine-grained semantic discrepancies between the image and target prompt.
We directly analyze how the input image $\vx$ aligns with the target prompt $y^{\text{tar}}$. Specifically, we identify the key entities in $\vx$ and $y^{\text{tar}}$:
\begin{equation}
\label{eq-10}
\begin{split}
\{y^i\}_{i=0}^n=\{y^0,y^1,...,y^n\} &\subseteq y^{\text{tar}},\\
\{e^i\}_{i=0}^m=\{e^0,e^1,...,e^m\} &\subseteq \text{Recaption}(\vx),
\end{split}
\end{equation} 
Then we utilize MLLMs (e.g., GPT4 \citep{openai2023gpt}, Gemini Pro \citep{team2023gemini}) to check the differences between $\{y^i\}_{i=0}^n$ and $\{e^i\}_{i=0}^m$ regarding numeric accuracy, attribute binding and object relationships. The resulting multimodal understanding feedback would be delivered to MLLMs for reason out editing plans.

\paragraph{CoT Planning for Editing}
Based on the captured semantic discrepancies between prompt and image, RPG triggers the CoT reasoning ability of MLLMs with high-quality filtered in-context examples, which involves manually designed step-by-step editing cases such as entity missing/redundancy, attribute mismatch, ambiguous relationships. Here, in our RPG, we introduce three main edit operations for dealing with these issues: addition $\text{Add}() $, deletion $\text{Del}()$,  modification $\text{Mod}()$. Take the multimodal feedback as the grounding context, RPG plans out a series of editing instructions.  An example $\text{Plan}(y^{\text{tar}},\vx)$ can be denoted as a composed operation list:
\begin{equation}
\begin{split}
     \label{eq-11}
    \{\text{Del}(y^{i},\vx),\cdots,\text{Add}(y^j,\vx),\cdots,
      \text{Mod}(y^k,\vx)\}, 
\end{split}
\end{equation}
where $i,j,k <= n, \text{length}(\text{Plan}(y^{\text{tar}},x^0)) = L$. In this way, we are able to decompose original complex editing task into simpler editing tasks for more accurate results.

\paragraph{Contour-based Regional Diffusion}
To collaborate more effectively with CoT-planned editing instructions, we generalize our complementary regional diffusion to text-guided editing. We locate and mask the target contour associated with the editing instruction \citep{kirillov2023segment}, and apply diffusion-based inpainting \citep{rombach2022high} to edit the target contour region according to the planned operation list $\text{Plan}(y^{\text{tar}},\vx)$. Compared to traditional methods that utilize cross-attention map swap or replacement \citep{hertz2022prompt,cao2023masactrl} for editing, our \textbf{mask-and-inpainting} method powered by CoT planning enables more accurate and complex editing operations (i.e., addition, deletion and  modification).

\paragraph{Multi-Round Editing for Closed-Loop Refinement}
Our text-guided image editing workflow is adaptable for a closed-loop self-refined text-to-image generation, which combines the contour-based editing with complementary regional diffusion generation.
We could conduct multi-round closed-loop RPG workflow controlled by MLLMs to progressively refine the generated image for aligning closely with the target text prompt. 
Considering the time efficiency, we set a maximum number of rounds to avoid being trapped in the closed-loop procedure. Based on this closed-loop paradigm, we can unify text-guided generation and editing in our RPG, providing more practical framework for the community.
\begin{table*}[ht]
\centering
\caption{Evaluation results on T2I-CompBench. 
RPG consistently demonstrates best performance regarding attribute binding, object relationships, and complex compositions.
We denote the best score in \colorbox{pearDark!20}{blue}, and the second-best score in \colorbox{mycolor_green}{green}. The baseline data is quoted from ~\citet{chen2023pixart}.} 
\label{benchmark:t2icompbench}
\resizebox{0.9\linewidth}{!}{ 
\begin{tabular}
{lcccccc}
\toprule
\multicolumn{1}{c}
{\multirow{2}{*}{\bf Model}} & \multicolumn{3}{c}{\bf Attribute Binding } & \multicolumn{2}{c}{\bf Object Relationship} & \multirow{2}{*}{\bf Complex$\uparrow$}
\\
\cmidrule(lr){2-4}\cmidrule(lr){5-6}

&
{\bf Color $\uparrow$ } &
{\bf Shape$\uparrow$} &
{\bf Texture$\uparrow$} &
{\bf Spatial$\uparrow$} &
{\bf Non-Spatial$\uparrow$} &
\\
\midrule
Stable Diffusion v1.4 \citep{rombach2022high}  & 0.3765 & 0.3576 & 0.4156 & 0.1246 & 0.3079 & 0.3080  \\
Stable Diffusion v2 \citep{rombach2022high}  & 0.5065 & 0.4221 & 0.4922 & 0.1342 & 0.3096 & 0.3386  \\
Composable Diffusion \citep{liu2022compositional} & 0.4063 & 0.3299 & 0.3645 & 0.0800 & 0.2980 & 0.2898  \\
Structured Diffusion \citep{feng2022training} & 0.4990 & 0.4218 & 0.4900 & 0.1386 & 0.3111 & 0.3355  \\
Attn-Exct v2 \citep{chefer2023attend} & 0.6400 & 0.4517 & 0.5963 & 0.1455 & 0.3109 & 0.3401  \\
GORS \citep{huang2023t2i}  & {0.6603} & 0.4785 & 0.6287 & 0.1815 & {0.3193} & 0.3328  \\
DALL-E 2 \citep{ramesh2022hierarchical} & 0.5750 & {0.5464} & {0.6374} & 0.1283 & 0.3043 & 0.3696  \\
SDXL \citep{betker2023improving} & 0.6369 & 0.5408 & 0.5637 & {0.2032} & 0.3110 & {0.4091}  \\

PixArt-$\alpha$ \citep{chen2023pixart} & {0.6886} & {0.5582} & \cellcolor{mycolor_green}{0.7044} & {0.2082} & {0.3179} & {0.4117}  \\
ConPreDiff \citep{yang2023improving} &\cellcolor{mycolor_green}{0.7019}&\cellcolor{mycolor_green}{0.5637}&{0.7021}&\cellcolor{mycolor_green}{0.2362}&\cellcolor{mycolor_green}{0.3195}&\cellcolor{mycolor_green}{0.4184}\\
\midrule
\textbf{RPG (Ours)}&\colorbox{pearDark!20}{0.8335}&\colorbox{pearDark!20}{0.6801}&\colorbox{pearDark!20}{0.8129}&\colorbox{pearDark!20}{0.4547}&\colorbox{pearDark!20}{0.3462}&\colorbox{pearDark!20}{0.5408}\\
\bottomrule
\end{tabular}
}
\vspace{-1em}
\end{table*}

\section{Experiments}

\subsection{Text-to-Image Generation}
\label{sec-exp-generation}
\paragraph{Implementation Details}
\label{detail}
Our RPG is general and extensible, we can incorporate \textbf{arbitrary} MLLM architectures and diffusion backbones \footnote{https://github.com/CompVis/stable-diffusion}\footnote{https://github.com/huggingface/diffusers}\footnote{https://github.com/hako-mikan/sd-webui-regional-prompter} into the framework. In our experiment, we choose GPT-4 \citep{openai2023gpt} as the recaptioner and CoT planner, and use SDXL \citep{podell2023sdxl} as the base diffusion backbone to build our RPG framework. 
Concretely, in order to trigger the CoT planning ability of MLLMs, we carefully design task-aware template and high-quality in-context examples to conduct few-shot prompting. 
\textbf{\textit{Base prompt}} and its weighted hyperparameter \textbf{\textit{base ratio}} are critical in our regional diffusion, we have provide further analysis in \cref{pic-RPG-base}. 
When the user prompt includes the entities with same class (e.g., two women, four boys), we need to set higher base ratio to highlight these distinct identities. 
On the contrary, when user prompt includes the the entities with different class name (e.g., ceramic vase and glass vase), we need lower base ratio to avoid the confusion between the base prompt and subprompts. 

\begin{figure}
\begin{center}\centerline{\includegraphics[width=1.\linewidth]{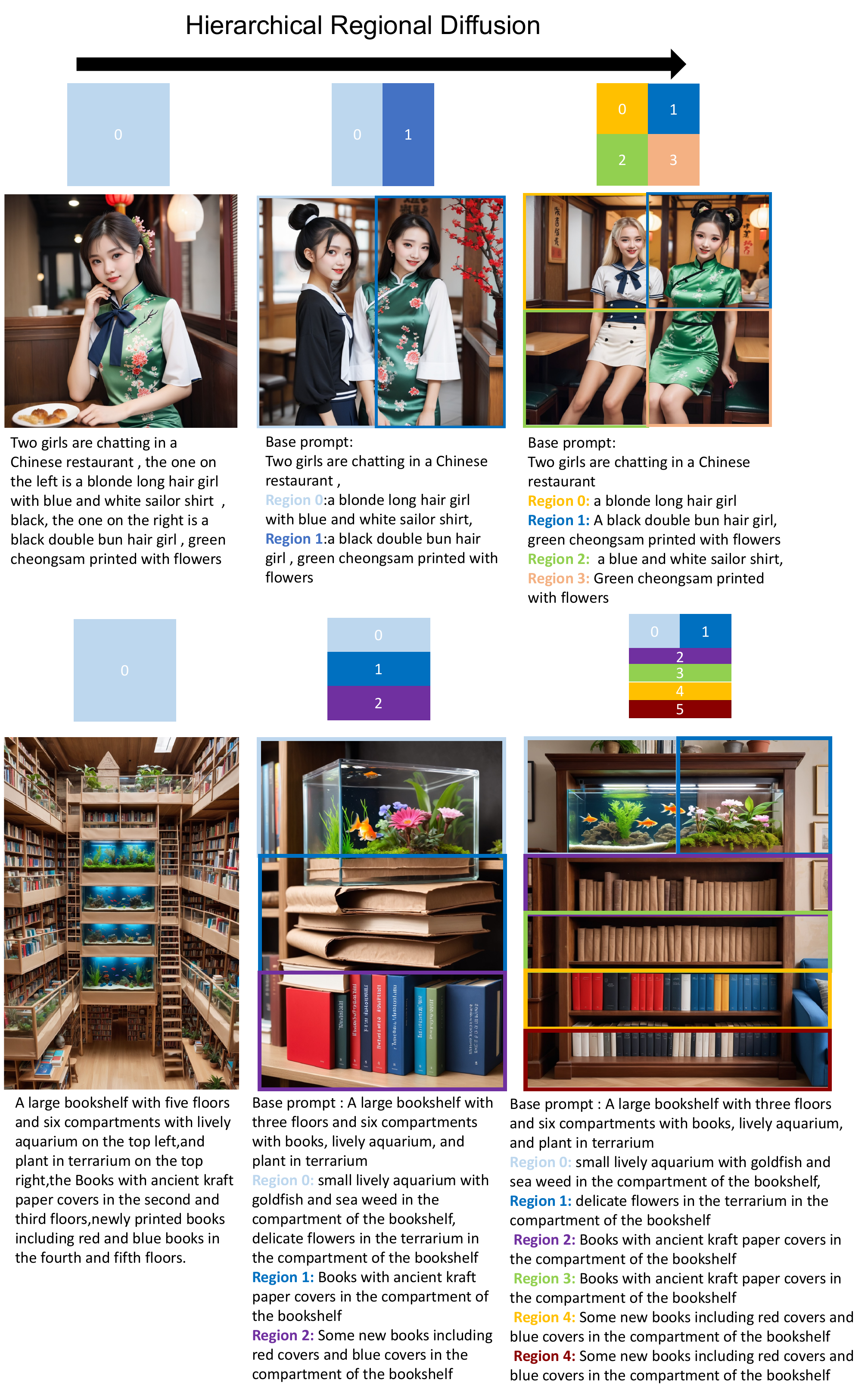}}
\caption{Demonstration of our hierarchical regional diffusion. Diffusion with more hierarchies can produce more satisfying results.}
\label{pic-RPG-hierarchical}
\end{center}
\vspace{-7mm}
\end{figure}

\begin{figure*}[ht]
\begin{center}\centerline{\includegraphics[width=1.\linewidth]{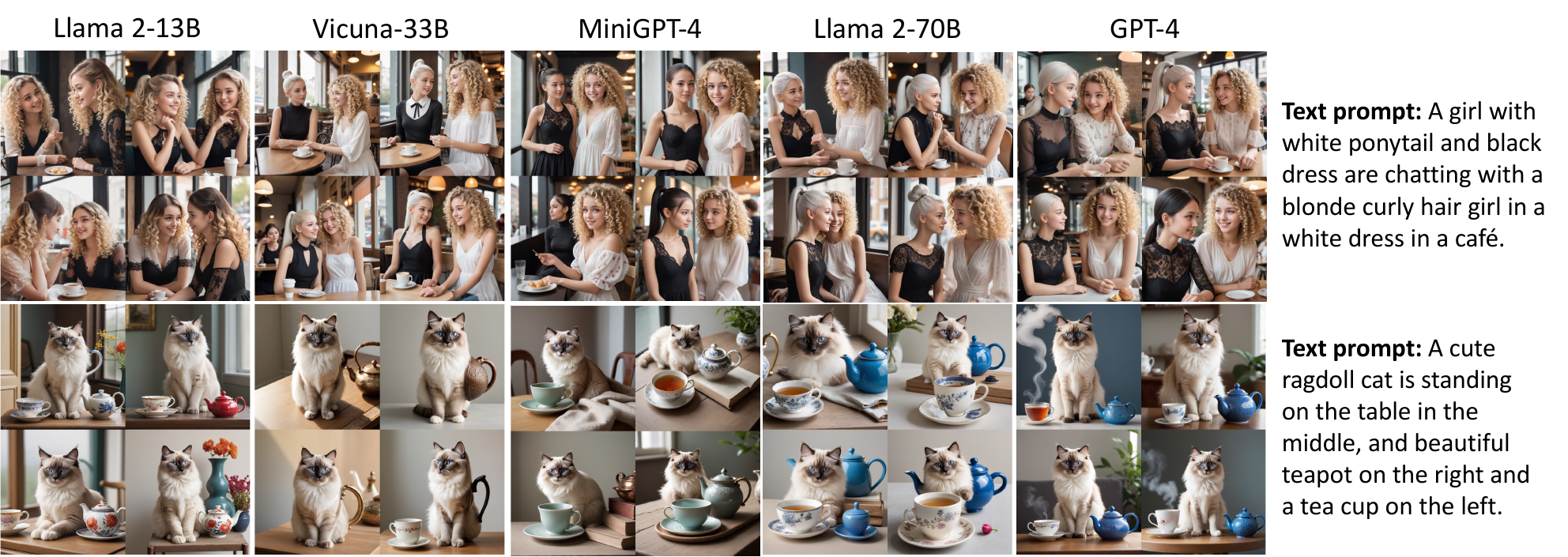}}
\vspace{-0.1in}
\caption{Generalizing RPG to different (multimodal) LLM architectures, including Llama 2 \citep{touvron2023llama2}, Vicuna \citep{chiang2023vicuna} and MiniGPT-4 \citep{zhu2023minigpt}.}
\label{pic-RPG-generalization-llm}
\end{center}
\vspace{-0.2in}
\end{figure*}

\paragraph{Main Results}
We compare with previous SOTA text-to-image models DALL-E 3 \citep{betker2023improving}, SDXL and LMD+ \citep{lian2023llm} in three main compositional scenarios: \textbf{(i) Attribute Binding}. Each text prompt in this scenario has multiple attributes that bind to different entities. \textbf{(ii) Numeric Accuracy}. Each text prompt in this scenario has multiple entities sharing the same class name, the number of each entity should be greater than or equal to two. \textbf{(iii) Complex Relationship}. Each text prompt in this scenario has multiple entities with different attributes and relationships (e.g., spatial and non-spational).  As demonstrated in \cref{benchmark:t2icompbench}, our RPG is significantly superior to previous models in all three scenarios, and achieves remarkable level of both fidelity and precision in aligning with text prompt. We observe that SDXL and DALL-E 3 have poor generation performance regarding numeric accuracy and complex relationship. In contrast, our RPG can effectively plan out precise number of subregions, and utilize proposed complementary regional diffusion to accomplish compositional generation. 
Compared to LMD+ \citep{lian2023llm}, a LLM-grounded layout-based text-to-image diffusion model, our RPG demonstrates both enhanced semantic expression capabilities and image fidelity. We attribute this to our CoT planning and complementary regional diffusion.
For quantitative results, we assess the text-image alignment of our method in a comprehensive benchmark, T2I-Compbench \citep{huang2023t2i}, which is utilized to evaluate the compositional text-to-image generation capability. 
In \cref{benchmark:t2icompbench}, we consistently achieve best performance among all methods proposed for both general text-to-image generation and compositional generation, including SOTA model ConPreDiff \citep{yang2023improving}.

\paragraph{Hierarchical Regional Diffusion}
We can extend our regional diffusion to a hierarchical format by splitting certain subregion to smaller subregions. As illustrated in \cref{pic-RPG-hierarchical}, when we increase the hierarchies of our region split, RPG can achieve a significant improvement in text-to-image generation. This promising result reveals that our complementary regional diffusion provides a new perspective for handling complex generation tasks and has the potential to generate arbitrarily compositional images.

\begin{figure}[ht]
\begin{center}\centerline{\includegraphics[width=0.85\linewidth]{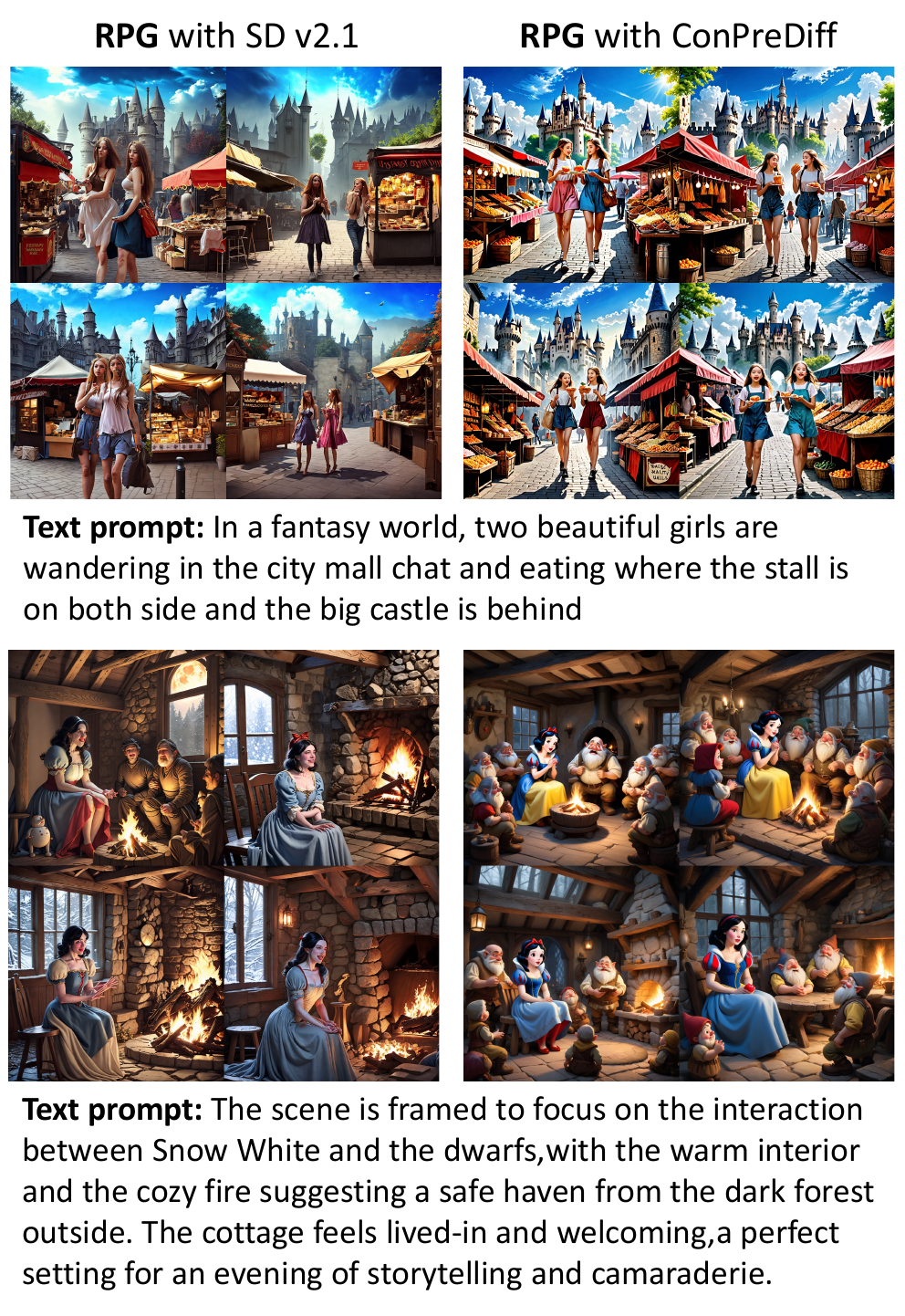}}
\caption{Generalizing RPG to different diffusion backbones, Stable Diffusion v2.1 \citep{rombach2022high} and recent SOTA diffusion model ConPreDiff \citep{yang2023improving}.}
\label{pic-RPG-generalization}
\end{center}
\vspace{-7mm}
\end{figure}

\paragraph{Generalizing to Various LLMs and Diffusion Backbones}
Our RPG framework is of great generalization ability, and can be easily generalized to various (M)LLM architectures (in \cref{pic-RPG-generalization-llm}) and diffusion backbones (in \cref{pic-RPG-generalization}). 
We observe that both LLM and diffusion architectures can influence the generation results.
We also generalize RPG to ControlNet \citep{zhang2023adding} for incorporating more conditional modalities. As demonstrated in \cref{pic-RPG-qualitative-controlnet}, our RPG can significantly improve the composibility of original ControlNet in both image fidelity and textual semantic alignment.

\begin{figure}[ht]
\begin{center}\centerline{\includegraphics[width=1\linewidth]{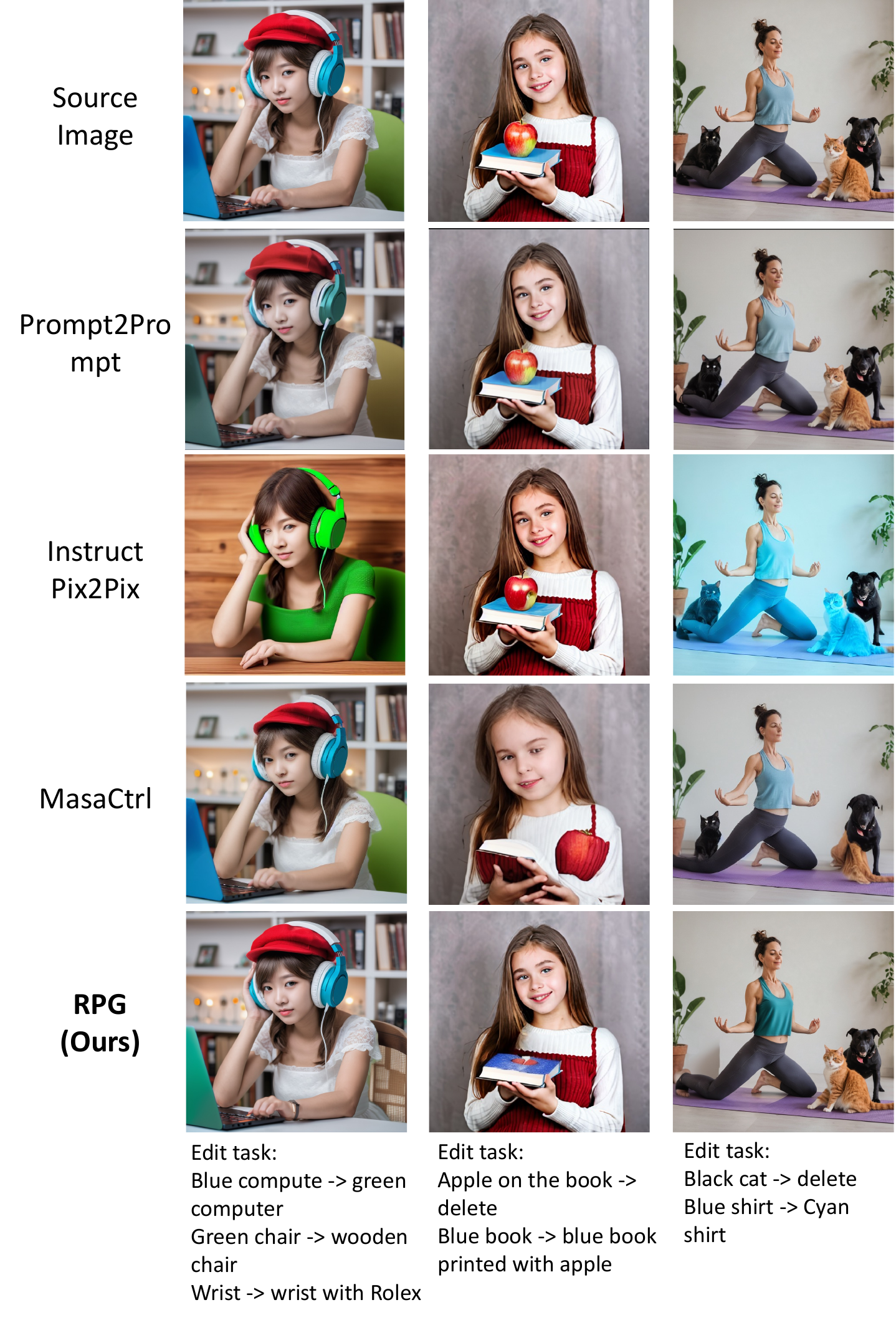}}
\caption{Qualitative comparison in text-guided image editing. We outperform previous powerful methods including Prompt2Prompt \citep{hertz2022prompt}, InstructPix2Pix \citep{brooks2023instructpix2pix} and MasaCtrl \citep{cao2023masactrl}.}
\label{pic-RPG-edit}
\end{center}
\vspace{-7mm}
\end{figure}

\subsection{Text-Guided Image Editing}
\paragraph{Qualitative Results}
In the qualitative comparison of text-guided image editing, we choose some strong baseline methods, including Prompt2Prompt \citep{hertz2022prompt}, InstructPix2Pix \citep{brooks2023instructpix2pix} and MasaCtrl \citep{cao2023masactrl}. Prompt2Prompt and MasaCtrl conduct editing mainly through text-grounded cross-attention swap or replacement, InstructPix2Pix aims to learn a model that can follow human instructions. As presented in \cref{pic-RPG-edit}, RPG produces more precise editing results than previous methods, and our mask-and-inpainting editing strategy can also perfectly preserve the semantic structure of source image.

\begin{figure}[t]
\begin{center}\centerline{\includegraphics[width=1.\linewidth]{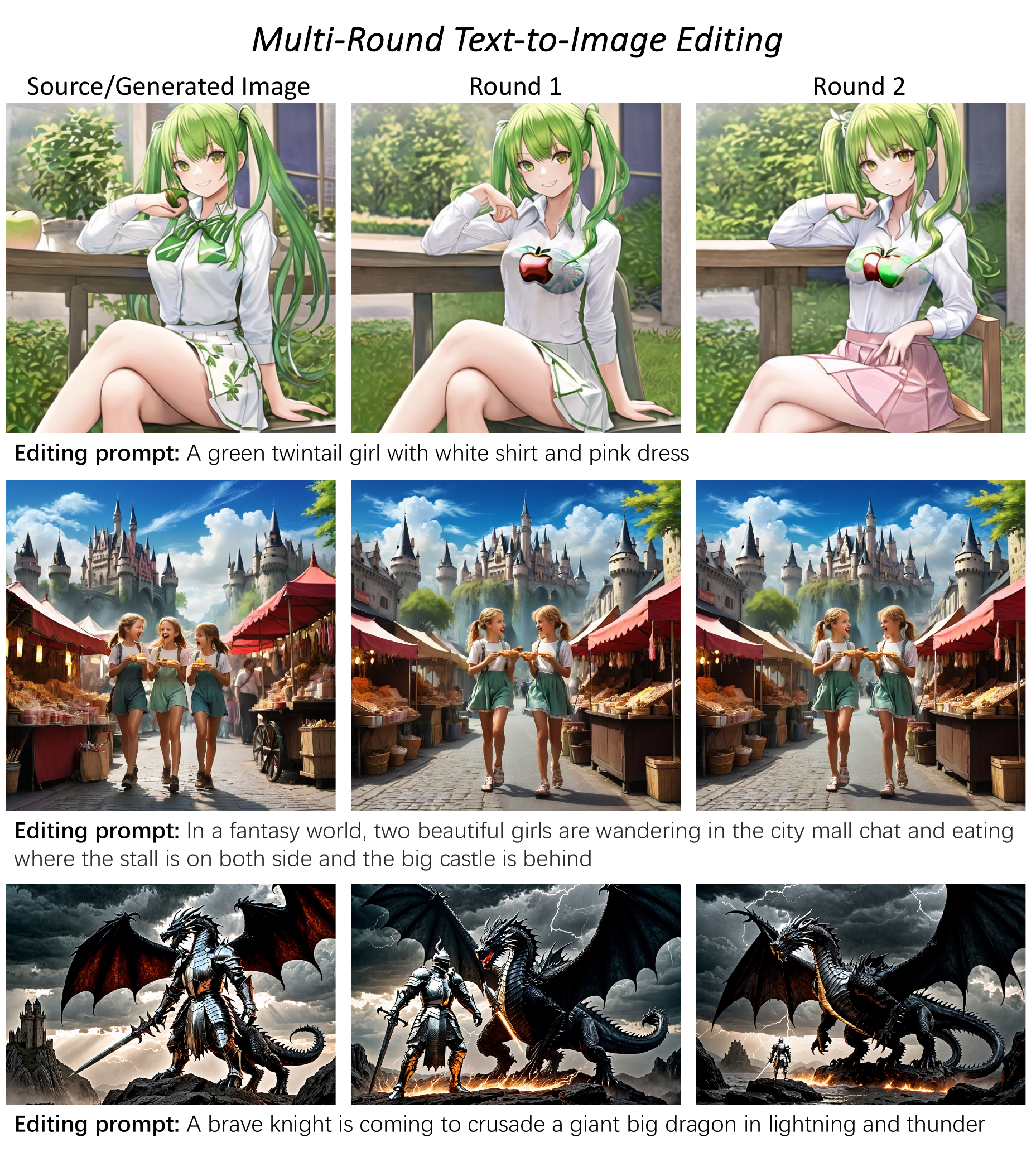}}
\caption{Multi-round text-guided image editing with our RPG framework.}
\label{pic-RPG-round}
\end{center}
\vspace{-7mm}
\end{figure}

\paragraph{Multi-Round Editing}
We conduct multi-round editing to evaluate the self-refinement with our RPG framework in \cref{pic-RPG-round}. We conclude that the self-refinement based on RPG can significantly improve precision, demonstrating the effectiveness of our recaptioning-based multimodal feedback and CoT planning. We also find that RPG is able to achieve satisfying editing results within 3 rounds.

\section{Model Analysis}
\label{sec-abaltion}
\begin{figure}[t]
\begin{center}\centerline{\includegraphics[width=1\linewidth]{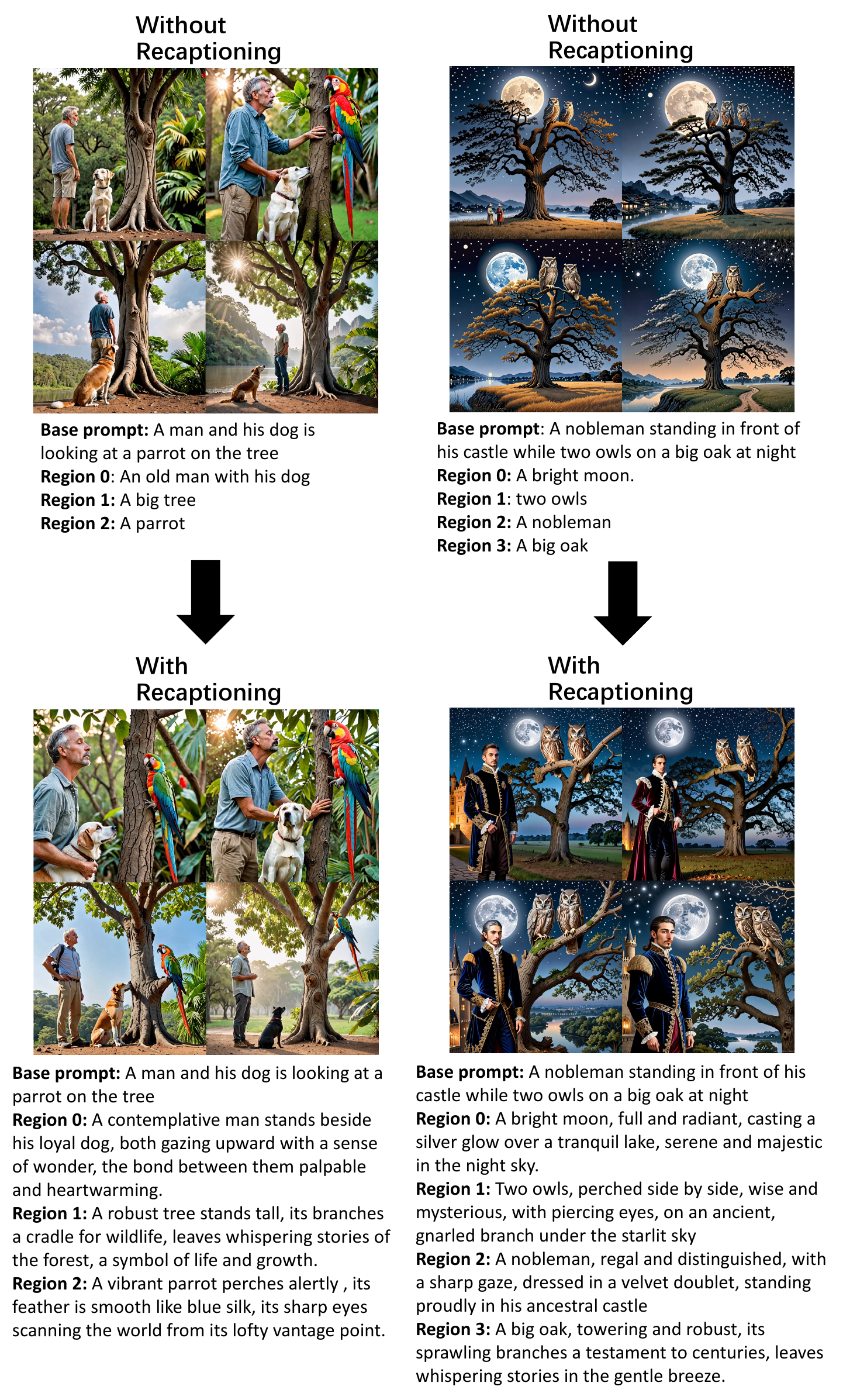}}
\caption{Ablation study of recaptioning in RPG.}
\label{pic-RPG-ablation-recap}
\end{center}
\vspace{-7mm}
\end{figure}

\paragraph{Effect of Recaptioning}
We conduct ablation study about the recaptioning, and show the result in \cref{pic-RPG-ablation-recap}. From the result, we observe that without recaptioning, the model tends to ignore some key words in the generated images. Our recaptioning can describe these key words with high-informative and denser details, thus generating more delicate and precise images.

\begin{figure}[t]
\begin{center}\centerline{\includegraphics[width=0.85\linewidth]{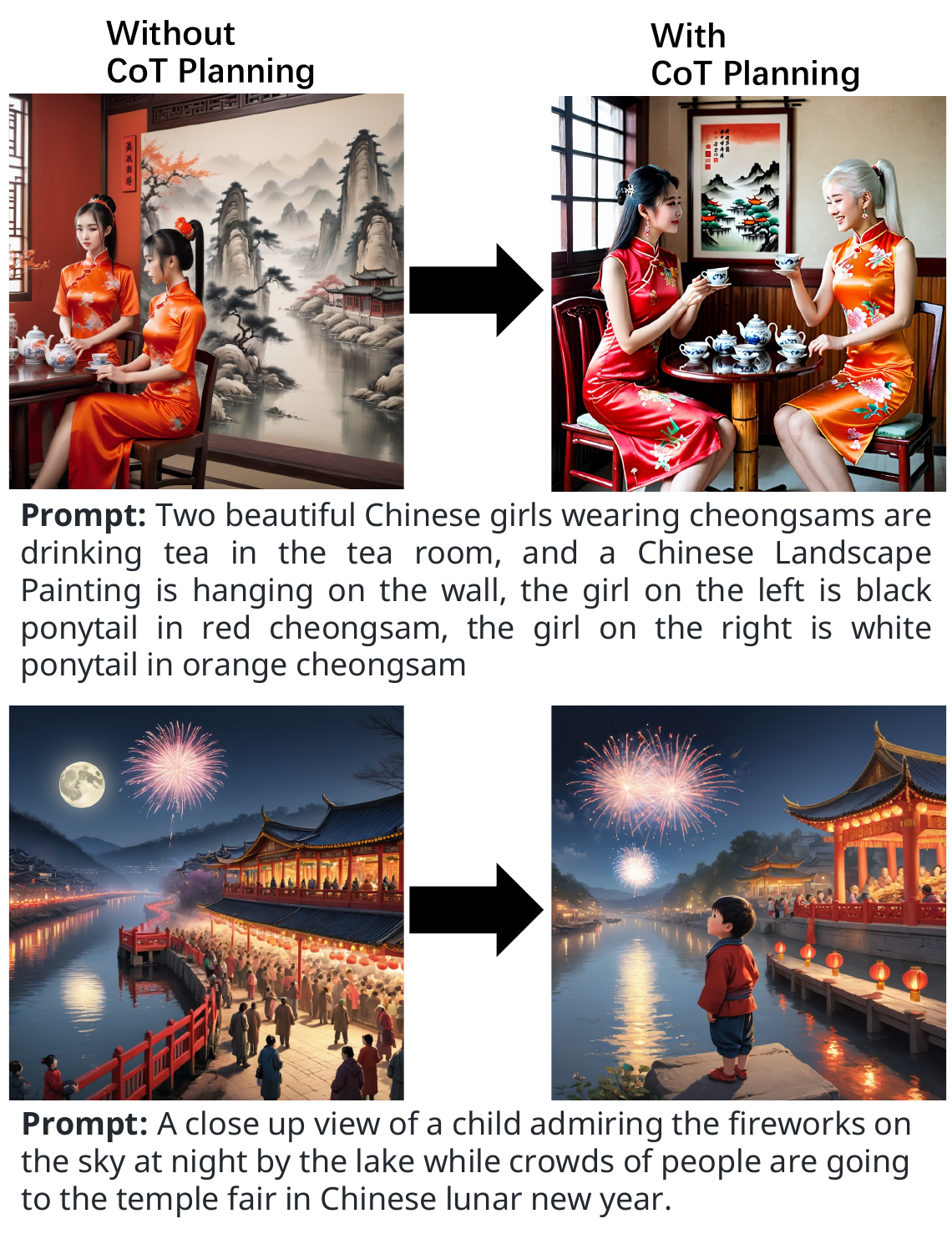}}
\caption{Ablation study of CoT planning in RPG.}
\label{pic-RPG-ablation-cot}
\end{center}
\vspace{-7mm}
\end{figure}

\paragraph{Effect of CoT Planning}
In the ablation study about CoT planning, as demonstrated in \cref{pic-RPG-ablation-cot}, we observe that the model without CoT planning fail to parse and convey complex relationships from text prompt. In contrast, our CoT planning can help the model better identify fine-grained attributes and relationships from text prompt, and express them through a more realistic planned composition.

\begin{figure}[t]
\begin{center}\centerline{\includegraphics[width=1.\linewidth]{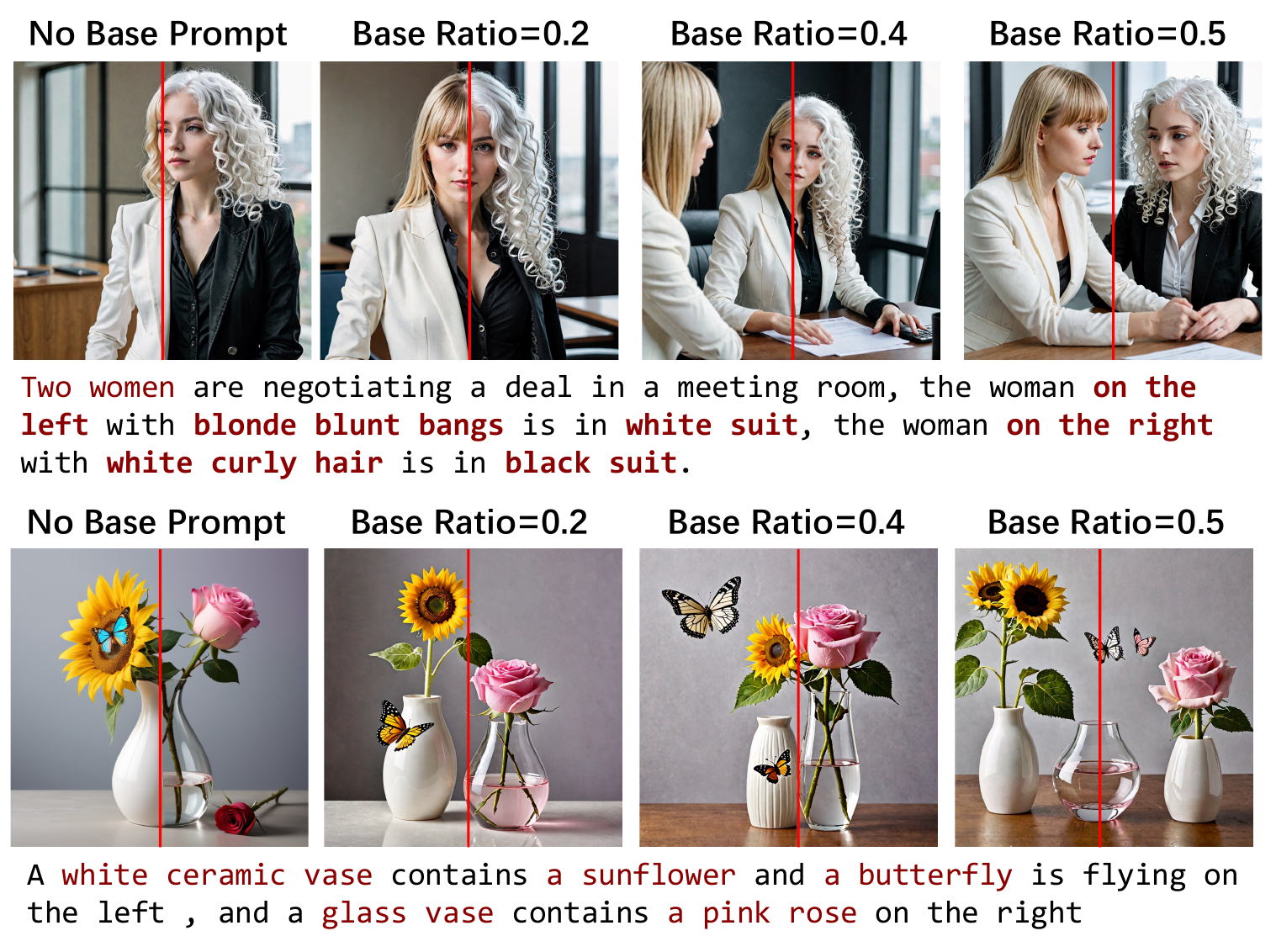}}
\caption{Ablation study of base prompt in complementary regional diffusion.}
\label{pic-RPG-base}
\end{center}
\vspace{-7mm}
\end{figure}

\paragraph{Effect of Base Prompt}
In RPG, we leverage the generated latent from base prompt in diffusion models to improve the coherence of image compositions. Here we conduct more analysis on it in \cref{pic-RPG-base}. From the results, we find that the proper ratio of base prompt can benefit the conjunction of different subregions, enabling more natural composition. Another finding is that excessive base ratio may result in undesirable results because of the confusion between the base prompt and regional prompt.

\section{Related Work}
\paragraph{Text-Guided Diffusion Models}
Diffusion models \citep{sohl2015deep,song2019generative,ho2020denoising,song2020improved,song2020score,yang2024structure} are a promising class of generative models, and \citet{dhariwal2021diffusion} have demonstrated the superior image synthesis quality of diffusion model over generative adversarial networks (GANs) \citep{reed2016generative,creswell2018generative}. GLIDE \citep{nichol2021glide} and Imagen \citep{saharia2022photorealistic} focus on the text-guided image synthesis, leveraging pre-trained CLIP model \citep{radford2021learning,raffel2020exploring} in the image sampling process to improve the  semantic alignment between text prompt and generated image. Latent Diffusion Models (LDMs) \citep{rombach2022high} move the diffusion process from pixel space to latent space for balancing algorithm efficiency and image quality. Recent advancements in text-to-image diffusion models , such as SDXL \citep{podell2023sdxl}, ContextDiff \citep{yang2024crossmodal} and DALL-E 3 \citep{betker2023improving}, further improve both quality and alignment from different perspectives. 
Despite their tremendous success, generating high-fidelity images with complex prompt is still challenging \citep{betker2023improving,huang2023t2i}. This problem is exacerbated when dealing with compositional descriptions involving spatial relationships, attribute binding and numeric awareness. In this paper, we aim to address this issue by incorporating the powerful CoT reasoning ability of MLLMs into text-to-image diffusion models.

\paragraph{Compositional Diffusion Generation} Recent researches aim to improve compositional ability of text-to-image diffusion models. Some approaches mainly introduce additional modules into diffusion models in training \citep{li2023gligen,avrahami2023spatext,zhang2023adding,mou2023t2i,yang2023reco,huang2023composer,huang2023t2i}. For example, GLIGEN \citep{li2023gligen} and ReCo \citep{yang2023reco} design position-aware adapters on top of the diffusion models for spatially-conditioned image generation. T2I-Adapter and ControlNet \citep{zhang2023adding,mou2023t2i} specify some high-level features of images for controlling semantic structures \citep{zhang2023controllable}. These methods, however, result in additional training and inference costs. Training-free methods aim to steer diffusion models through manipulating latent or cross-attention maps according to spatial or semantic constraints during inference stages \citep{feng2022training,liu2022compositional,hertz2022prompt,cao2023masactrl,chen2024training,chefer2023attend}. Composable Diffusion \citep{liu2022compositional} decomposes a compositional prompt into smaller sub-prompts to generate distinct latents and combines them with a score function. \citet{chen2024training} and \citet{lian2023llm} utilize the bounding boxes (layouts) to propagate gradients back to the latent and enable the model to manipulate the cross-attention maps towards specific regions.
Other methods apply Gaussian kernels \citep{chefer2023attend} or incorporate linguistic features \citep{feng2022training,rassin2023linguistic} to manipulate the cross-attention maps. 
Nevertheless, such manipulation-based methods can only make rough controls, and often lead to unsatisfied compositional generation results, especially when dealing with overlapped objects \citep{lian2023llm,cao2023masactrl}. Hence, we introduce an effective \textit{training-free complementary regional diffusion model}, grounded by MLLMs, to progressively refine image compositions with more precise control in the sampling process.

\paragraph{Multimodal LLMs for Image Generation}
Large Language Models (LLMs) \citep{chatgpt2022introducing,chung2022scaling,zhang2022opt,iyer2022opt,workshop2022bloom,muennighoff2022crosslingual,zeng2022glm,taylor2022galactica,chowdhery2023palm,chen2023llava,zhu2023minigpt,touvron2023llama,yang2023baichuan,li2023blip} have profoundly impacted the AI community. Leading examples like ChatGPT \citep{chatgpt2022introducing} have showcased the advanced language comprehension and reasoning skills through techniques such as instruction tuning \citep{ouyang2022training,li2023stablellava,zhang2023enhanced,liu2023visual}. Further, Multimodal Large language Models (MLLMs),  \citep{koh2023generating,yu2023scaling,sun2023generative,dong2023dreamllm,fu2023guiding,pan2023kosmos,wu2023visual,zou2023generalized,yang2023mm,gupta2023visual,suris2023vipergpt} integrate LLMs with vision models to extend their impressive abilities from language tasks to vision tasks, including image understanding, reasoning and synthesis. The collaboration between LLMs \citep{chatgpt2022introducing,openai2023gpt} and diffusion models \citep{ramesh2022hierarchical,betker2023improving} can significantly improve the text-image alignment as well as the quality of generated images \citep{yu2023scaling,chen2023llava,dong2023dreamllm,wu2023self,feng2023layoutgpt,pan2023kosmos}. For instance, GILL \citep{koh2023generating} can condition on arbitrarily interleaved image and text inputs
to synthesize coherent image outputs, and Emu \citep{sun2023generative} stands out as a generalist
multimodal interface for both image-to-text and text-to-image tasks.
Recently, LMD \citep{lian2023llm} utilizes LLMs to enhance the compositional generation of diffusion models by generating images grounded on bounding box layouts from the LLM \citep{li2023gligen}.
However, existing works mainly incorporate the LLM as a simple plug-in component into diffusion models, or simply take the LLM as a layout generator to control image compositions.  
In contrast, we utilize MLLMs to plan out image compositions for diffusion models where MLLMs serves as a global task planner in both \textit{region-based} generation and editing process.
\section{Conclusion}
In this paper, aiming to address the challenges of complex or compositional text-to-image generation, we propose a SOTA training-free framework RPG, harnessing MLLMs to master diffusion models. In RPG, we propose complementary regional diffusion models to collaborate with our designed MLLM-based recaptioner and planner. Furthermore, our RPG can unify text-guided imgae generation and editing in a closed-loop approach, and is capable of generalizing to any MLLM architectures and diffusion backbones. For future work, we will continue to improve this new framework for incorporating more complex modalities as input condition, and extend it to more realistic applications.

\bibliography{main}
\bibliographystyle{icml2024}




\end{document}